%% file: barl_2020.tex
\documentclass{article}

\usepackage[final]{neurips_2020}

\usepackage[utf8]{inputenc} % allow utf-8 input
\usepackage[T1]{fontenc}    % use 8-bit T1 fonts
\usepackage[usenames, dvipsnames]{color} % Cool colors
\usepackage{enumerate, amsmath, amsthm, amssymb, algpseudocode, pifont, subfig, fullpage, csquotes, dashrule, tikz, bbm, booktabs, bm, verbatim, url}
\usepackage[framemethod=TikZ]{mdframed}
\usepackage{natbib}
\usepackage[ruled,vlined]{algorithm2e}
\usepackage[colorlinks]{hyperref}
\hypersetup{
     colorlinks = true,
     linkcolor = blue,
     anchorcolor = blue,
     citecolor = red,
     filecolor = blue,
     urlcolor = blue
     }

\input{commands}

\newif\ifsubmit
\submitfalse
\ifsubmit
\newcommand{\dnote}[1]{}
\newcommand{\pnote}[1]{}
\newcommand{\plnote}[1]{}
\else
\newcommand{\dnote}[1]{\textcolor{blue}{Dilip: #1}}
\newcommand{\pnote}[1]{\textcolor{orange}{Peter: #1}}
\newcommand{\plnote}[1]{\textcolor{green}{Pierre-Luc: #1}}
\fi
\title{An Information-Theoretic Perspective on \\ Credit Assignment in Reinforcement Learning}

\author{%
  Dilip Arumugam \\
  Department of Computer Science\\
  Stanford University\\
  \texttt{dilip@cs.stanford.edu} \\
   \And
  Peter Henderson \\
  Department of Computer Science\\
  Stanford University\\
  \texttt{phend@cs.stanford.edu} \\
  \And
  Pierre-Luc Bacon \\
  Mila - University of Montreal\\
  \texttt{pierre-luc.bacon@mila.quebec} \\
}

\begin{document}

\maketitle

\begin{abstract}  
    How do we formalize the challenge of credit assignment in reinforcement learning? 
    Common intuition would draw attention to reward sparsity as a key contributor to difficult credit assignment and traditional heuristics would look to temporal recency for the solution, calling upon the classic eligibility trace.
     We posit that it is not the sparsity of the reward itself that causes difficulty in credit assignment, but rather the \emph{information sparsity}. We propose to use information theory to define this notion, which we then use to characterize when credit assignment is an obstacle to efficient learning. 
    With this perspective, we outline several information-theoretic mechanisms for measuring credit under a fixed behavior policy, highlighting the potential of information theory as a key tool towards provably-efficient credit assignment.
\end{abstract}

\section{Introduction}

The \textit{credit assignment} problem in reinforcement learning~\citep{minsky1961steps,sutton1985temporal,sutton1988learning} is concerned with identifying the contribution of past actions on observed future outcomes. Of particular interest to the reinforcement-learning (RL) problem~\citep{sutton1998introduction} are observed future returns and the value function, which quantitatively answers \textit{``how does choosing an action $a$ in state $s$ affect future return?''} Indeed, given the challenge of sample-efficient RL in long-horizon, sparse-reward tasks, many approaches have been developed to help alleviate the burdens of credit assignment~\citep{sutton1985temporal,sutton1988learning,sutton1998introduction,singh1996reinforcement,precup2000eligibility,riedmiller2018learning,harutyunyan2019hindsight,hung2019optimizing,arjona2019rudder,ferret2019credit,trott2019keeping,van2020expected}. 

The long-horizon, sparse-reward problems examined by many existing works on efficient credit assignment are often recognized as tasks that require prolonged periods of interaction prior to observing non-zero feedback; in the extreme case, these are ``goal-based'' problems with only a positive reward at terminal states and zero rewards everywhere else.
To see why the sparsity of rewards cannot serve as a true hardness measure of credit assignment in RL, consider any sparse-reward MDP. Notice that for any fixed constant $c > 0$, we can have a RL agent interact with the same MDP except under the reward function $\tilde{\mc{R}}(s,a) = \mc{R}(s,a) + c$. Clearly, this new reward function is no longer sparse; in fact, it is guaranteed to offer non-zero feedback on every timestep. And yet, it is also clear that this modification has neither changed the optimal policy $\pi^\star$ nor has it alleviated any of the burdens of credit assignment that stand in the way of efficiently learning $\pi^\star$. This simple example illustrates how rectifying the sparsity of rewards does not yield a reduction in the difficulty of credit assignment. And yet, there are well-known examples of how a prudent choice of reward bonus can substantially accelerate learning~\citep{ng1999policy}. 

While it seems rather easy to show that the sparsity of reward is not the key factor that determines the hardness of credit assignment, the intuitive connection between reward sparsity and the difficulty of credit assignment persists; several works decompose problem difficulties into dichotomies of sparse- and dense-reward tasks~\citep{romoff2019separating,bellemare2013arcade}. In this work, we maintain that while sparse-reward problems may serve as quintessential examples of decision-making problems where credit assignment is challenging, the underlying mechanism that drives this hardness can be more aptly characterized using information theory. We make this precise by introducing \textit{information sparsity} as a formalization of the real driving force behind the credit assignment challenge; namely, a lack of information between behavior (actions taken under a particular behavior policy) and observed returns, yielding a case of information scarcity. 

Beyond clarifying what makes credit assignment difficult, our goal is to show that information theory can also serve as a tool for facilitating efficient credit assignment. To that end, we offer several information-theoretic measures with which an agent may quantitatively allocate credit to specific state-action pairs, under a fixed behavior policy. Each of our proposed measures quantifies a precise relationship between behavior and trajectory returns. We then expand our consideration to not just the single full return of a trajectory, but the entire sequence of returns encountered at each state-action pair, exposing a connection with causal information theory. Our work leaves open the question of how these information-theoretic connections with credit assignment may integrate into existing RL algorithms. More broadly, we hope that the insights presented here inspire subsequent research on the role of information theory in analyzing efficient credit assignment in RL. 

Our paper proceeds as follows: we define our problem formulation in Section \ref{sec:form}, introduce our notion of information sparsity in Section \ref{sec:info_sparse}, outline further directions for information-theoretic credit assignment in Section \ref{sec:info_ca}, and conclude with discussions of future work in Section \ref{sec:disc}. Due to space constraints, background on information theory, related work, and all proofs have been relegated to the appendix.

\section{Problem Formulation}
\label{sec:form}

We consider a finite-horizon Markov Decision Process (MDP)~\citep{bellman1957markovian,Puterman94} defined by $\mc{M} = \langle \mc{S}, \mc{A}, \mc{R}, \mc{T}, H, \beta, \gamma \rangle$ where $\mc{S}$ denotes the state space, $\mc{A}$ is the action set, $\mc{R}:\mc{S} \times \mc{A} \mapsto \bR$ is a (deterministic) reward function, $\mc{T}:\mc{S} \times \mc{A} \mapsto \Delta(\mc{S})$ is the transition function producing a distribution over next states given the current state-action pair, $H$ is the finite, fixed episode horizon, $\beta \in \Delta(\mc{S})$ is the initial state distribution, and $\gamma \in [0,1)$ is the discount factor. We assume that both $\mc{S}$ and $\mc{A}$ are finite and use $|\mc{S}| = S$ and $|\mc{A}| = A$ to denote their respective sizes. At each timestep $h \in [H]$ of the current episode, the agent observes the current state $s_h$ and samples an action $a_h$ according to its current (stochastic) policy $\pi_h:\mc{S} \mapsto \Delta(\mc{A})$. The agent's objective is to find a policy so as to maximize the expected sum of future discounted rewards $\bE[\sum\limits_{h=1}^H \gamma^{h-1}\mc{R}(s_h,a_h)]$, where the expectation is taken with respect to randomness in the initial state, environment transitions, and policy. 
The value function of a (non-stationary) policy $\pi = (\pi_1,\ldots,\pi_H)$ denotes the expected future return by following the policy from a given state $s$ at timestep $h$,
$V_h^\pi(s) = \bE[\sum\limits_{h'=h}^H \gamma^{h'-h}\mc{R}(s_{h'}, a_{h'}) | s_h = s]$,
where the expectation is taken with respect to the policy $\pi$ and the environment transition dynamics $\mc{T}$. Similarly, we use the Bellman equation to define the action-value function $Q^\pi(s,a)$ representing the expected future return from timestep $h$, taking action $a$ from state $s$, and following policy $\pi$ thereafter, $Q_h^\pi(s, a) = \mc{R}(s,a) + \gamma\bE_{s' \sim \mc{T}(\cdot | s, a)}[V_{h+1}^\pi(s')]$, where $V_{H+1}^\pi(s) = 0$.

A fixed behavior policy $\pi$ induces a stationary visitation distribution over states and state-action pairs denoted as $d^\pi(s)$ and $d^\pi(s,a)$, respectively. Moreover, we let $\rho^\pi(\tau)$ denote the distribution over trajectories generated by a policy $\pi$ with $\rho^\pi(\tau | s)$ and $\rho^\pi(\tau | s,a)$ conditioning on a particular choice of start state $s$ or starting state-action pair $(s,a)$ respectively. Following the notation from distributional RL~\citep{bellemare2017distributional}, we let $Z(\tau)$ be a random variable denoting the random return obtained after completing trajectory $\tau \sim \rho^\pi(\cdot)$ under behavior policy $\pi$; analogously, $Z \triangleq Z(s,a)$ is a random variable denoting the return observed at state $s$ having taken $a$ and then following $\pi$ thereafter\footnote{For clarity, we ignore issues that arise from the mismatch between the continuous random variables $Z(\tau),Z$ and discrete variables involving states, actions, and trajectories. Instead, we think of returns as discrete random variables obtained from a sufficiently fine quantization of the real-valued return.}. Given a trajectory $\tau = \{(s_1,a_1),(s_2,a_2),\ldots,(s_{H-1},a_{H-1}),(s_H,a_H) \}$, we may ``index'' into its constituent state-action pairs using the following notation: $\tau_h = (s_h,a_h)$, $\tau^h = \{(s_1,a_1),(s_2,a_2),\ldots,(s_h,a_h)\}$, $\tau_i^j = \{(s_i,a_i),(s_{i+1},a_{i+1}),\ldots,(s_{j-1},a_{j-1}),(s_j,a_j)\}$, and $\tau^{-h} = \{(s_1,a_1),\ldots,(s_{h-1},a_{h-1}), (s_{h+1},a_{h+1}),\ldots,(s_H,a_H)\}$.

\section{Information Sparsity}
\label{sec:info_sparse}

The sparsity of rewards is a property of MDPs often mentioned when describing ``difficult'' decision-making problems. While most works offer a verbal explanation of what constitutes a sparse reward MDP, few papers~\citep{riedmiller2018learning,trott2019keeping} offer a precise definition through the specification of the reward function as $\mc{R}(s,a,s') = \delta_{s_g}(s')$ if  $d(s',s_g) \leq \epsilon$ and $\mc{R}(s,a,s') = 0$ otherwise, where $d: \mc{S} \times \mc{S} \ra \bR_{\geq 0}$ is a metric on the state space, $\epsilon$ is a small constant, and $\delta_{s_g}(s')$ is an arbitrary function defining the reward structure for states within $\epsilon$ distance of some goal state $s_g$, as measured by $d$; the common choice is to have $\delta_{s_g}(s')$ be a constant function (for instance, $\delta_{s_g}(s') = 1$).

While a complete lack of feedback across several timesteps demands judicious exploration to encounter the first nontrivial reward signal, exploration is not the only complication. Even after an agent has acquired the first non-zero reward, it still faces a demanding credit-assignment challenge wherein it must decide which step(s) of a long trajectory were critical to the observed outcome. Stemming from this fact, reward sparsity and the credit-assignment challenge are often discussed together leading to a notion that the former is itself a driving force behind the difficulty of the latter.

In this work, we maintain that this phenomenon can be explained via information theory. Recalling the example posed in the introduction, we call attention to how the addition of a positive constant to a sparse reward function, while eliminating sparsity, offers no useful information. In contrast, a more careful choice of, for example, negated distance to goal removes sparsity in an informative way. We can make this precise by examining the following quantity:
\begin{align}
    \mc{I}^{\pi}_{s,a}(Z) = \kl{p(Z | s, a)}{p(Z|s)},
    \label{eq:info_sparse_kl}
\end{align}

where $p(Z|s,a)$ denotes the distribution over returns for a random state-action pair conditioned on a particular realization of the state and action. Analogously, $p(Z|s) = \sum\limits_{a \in \mc{A}} \pi(a|s)p(Z|s, a)$ denotes the distribution over the random returns for the state-action pair conditioned on a particular realization of the state. The quantity $\mc{I}^{\pi}_{s,a}(Z)$ is itself a random variable depending on the particular realization of the state-action pair $(s,a)$. 

Intuitively, Equation \ref{eq:info_sparse_kl} measures how much the distribution over returns of a given state-action pair shifts relative to the distribution over returns for the particular state, marginalizing over all actions. Recalling that $Q^\pi(s,a) = \bE_{p(Z|s, a)}[Z]$ and $V^\pi(s) = \bE_{p(Z|s)}[Z]$, one may interpret Equation \ref{eq:info_sparse_kl} as distributional analogue to the advantage function $A^\pi(s,a) = Q^\pi(s,a) - V^\pi(s)$. To connect this quantity with information theory, we need only apply an expectation:
\begin{align}
    \mc{I}(A;Z|S) &= \bE_{(s,a) \sim d^\pi}\left[\kl{p(Z | s, a)}{p(Z|s)}\right]
    \label{eq:info_sparse_mi}
\end{align}

This quantity carries a very intuitive meaning in the context of credit assignment: conditioned upon states visited by policy $\pi$, how much information do the actions of $\pi$ carry about the returns of those state-action pairs? Difficulties with overcoming the credit-assignment challenge in long-horizon problems arise when $\mc{I}(A;Z|S)$ is prohibitively small. That is, a decision-making problem where the actions of the initial policy have little to no dependence on returns cannot acquire the signal needed to learn an optimal policy; sparse-reward problems are a natural example of this. More formally, we can define this notion of \textit{information sparsity} as follows:
\newpage
 
 \ddef{Information Sparsity}{
Given an MDP $\mc{M}$ with non-stationary policy class $\Pi^H$, let $\Pi_0 \subset \Pi^H$ denote the set of initial policies employed at the very beginning of learning in $\mc{M}$. For a small constant $\varepsilon > 0$, we classify $\mc{M}$ as $\varepsilon$-information-sparse if
\begin{align*}
    \sup\limits_{\pi_0 \in \Pi_0} \mc{I}^{\pi_0}(A;Z|S) \leq \varepsilon
\end{align*}
\label{def:info_sparse}
}

Under Definition 1, $\varepsilon$-sparse MDPs with small parameter $\varepsilon$ (inducing higher sparsity of information) represent a formidable credit-assignment challenge. Using sparse reward problems as an illustrative example and taking information sparsity to be the core obstacle to efficient credit assignment, we may consider how various approaches to dealing with such tasks also resolve information sparsity. Perhaps the most common approach is to employ some form of intrinsic motivation or reward shaping~\citep{ng1999policy,chentanez2005intrinsically} with heuristics such as the distance to goal, curiosity, or random network distillation~\citep{pathak2017curiosity,burda2018exploration}. In all of these cases, reward sparsity is resolved in a manner that also corrects for information sparsity; to help visualize this, consider a sufficiently-large gridworld MDP with actions for each cardinal direction and a goal-based reward function. Prior to reward augmentation, sparse rewards would likely result in returns of zero across the entire space of state-action pairs visited by a uniform random policy. In contrast, by using a reward bonus equal to, for instance, the negated distance to goal, individual actions taken in almost every state can create meaningful deviations between the distributions $p(Z|s,a)$ and $p(Z|s)$, translating into an increase in the available bits of information measured by information sparsity. A similar comment can also be made for approaches that invoke Thompson sampling~\citep{thompson1933likelihood} as a tool for facilitating deep exploration~\citep{osband2016deep,osband2019deep}; the random noise perturbations used by such approaches translate into excess information that accumulates in the $\mc{I}^{\pi_0}(A;Z|S)$ term.

Alternatively, there are other techniques for handling credit assignment that either change the problem formulation altogether or address the long-horizon aspect of decision making. In the latter category, the options framework~\citep{sutton1999between,bacon2017option} has served as a powerful tool for accommodating efficient RL over long horizons by adopting a two-timescale approach. Similar to a judicious choice of reward shaping function, provision of the right options to an agent can eliminate the difficulty of credit assignment that stems from having a long horizon. In our framework, this can be seen as picking a new (hierarchical) policy class $\Pi_0$ to resolve information sparsity. Finally, some approaches simply shift to the multi-task setting and assume access to a function that can identify failed trajectories of one task as successful behaviors for other tasks~\citep{andrychowicz2017hindsight}. As long as these hindsight approaches can generate informative feedback for some subset of the task distribution, they can bootstrap learning of more complicated tasks.

To conclude this section, we consider the computability of information sparsity and recall that, for any function $f: \mc{S} \times \mc{A} \ra \bR$,
\begin{align}
    \bE_{(s,a) \sim d^\pi}[f(s,a)] &= \bE_{\tau \sim \rho^\pi}[\sum\limits_{h=1}^H \gamma^{h-1} f(s_h,a_h)]
\end{align}
Taking $f(s,a) = \kl{p(Z | s, a)}{p(Z|s)}$, it follows that we need only choose an algorithm for recovering $p(Z|s,a)$~\citep{morimura2010nonparametric,morimura2012parametric,bellemare2017distributional} to compute information sparsity.

\section{Information-Theoretic Credit Assignment}
\label{sec:info_ca}

In this section, we present potential quantities of interest for deciding how to award credit to an individual state-action pair given the outcome (return) of an entire trajectory.

\subsection{Measuring Credit}

One possible choice for deciding how responsible or culpable a single step of behavior is for the outcome of the whole trajectory is by performing a sort of sensitivity analysis wherein a single point in the trajectory is varied while all other points are held fixed. This first proposition for measuring the credit of state-action pairs embodies this idea exactly using conditional mutual information.

\begin{proposition}
Let $\pi$ be a fixed behavior policy such that $\tau \sim \rho^\pi$. Let $R_h$ be a random variable denoting the reward observed at timestep $h$ (where the randomness of the deterministic reward follows from the randomness of the state-action pair at $h$, $\tau_h$). It follows that:
\begin{align*}
    \mc{I}\left(Z(\tau);\tau_h|\tau^{-h}\right) &= \mc{H}\left(R_h | \tau^{h-1}\right)
\end{align*}
\label{prop:credit}

The proof is provided in Appendix~\ref{sec:proofs}.
\end{proposition}

The left-hand side of Proposition \ref{prop:credit} is a conditional mutual information term quantifying the information between a single state-action pair $\tau_h$ and the policy return $Z(\tau)$, conditioned on the entire trajectory excluding timestep $h$. The statement of Proposition \ref{prop:credit} shows that this measure of credit is equal to the entropy in rewards conditioned on the trajectory up to timestep $h-1$, $\tau^{h-1}$. In practice, this encourages an approach that is reminiscent of RUDDER~\citep{arjona2019rudder} wherein a recurrent neural network learns a representation of $\tau^{h-1}$ and is trained as a reward classifier (for some discretization of the reward interval); the entropy of the resulting classifier can then be used as a weighting strategy for policy parameter updates or to bias exploration as a reward bonus.

Alternatively, it may be desirable to examine the importance of the current state-action pair towards policy returns conditioned only on the past history, $\tau^{h-1}$. To help facilitate such a measure, it is useful to recall the hindsight distribution $h(a|s,Z(\tau)$ of \citet{harutyunyan2019hindsight} that captures the probability of having taken action $a$ from state $s$ conditioned on the observed trajectory return $Z(\tau)$. 
\begin{proposition}
Let $\pi$ be a fixed behavior policy such that $\tau \sim \rho^\pi$ and let $h(a|s,Z(\tau))$ be the hindsight distribution as defined above. We have that
\begin{align*}
    \mc{I}(Z(\tau);\tau_h|\tau^{h-1}) &= \bE_{\tau^h}\left[\bE_{Z(\tau) | \tau^h}\left[ \log\left(\frac{h(a_h|s_h, Z(\tau))}{\pi(a_h | s_h)}\right)\right] \right]
\end{align*}

Moreover,
\begin{align*}
    \mc{I}(Z(\tau);\tau) &= \sum\limits_{h=1}^H \bE_{\tau^h}\left[\bE_{Z(\tau) | \tau^h}\left[ \log\left(\frac{h(a_h|s_h, Z(\tau))}{\pi(a_h | s_h)}\right)\right] \right]
\end{align*}
\label{prop:hca}
The proof is provided in Appendix~\ref{sec:hingsight_proof}.
\end{proposition}

Proposition \ref{prop:hca} tells us that learning the hindsight distribution as proposed in \citet{harutyunyan2019hindsight} can also be effectively used to tackle credit assignment in an information-theoretic manner by estimating the conditional mutual information between individual state-action pairs and returns, conditioned on the trajectory up to the previous timestep. While this measure captures a useful quantity intuitively, how to best incorporate such an estimate into an existing RL algorithm remains an open question for future work.

\subsection{Causal Information Theory \& Hindsight}

In the previous section, we examined the information content between a trajectory and its return, leveraging the fact that the trajectory random variable is a sequence of random variables denoting the individual state-action pairs. In this section, we draw attention to the fact that individual returns, like the state-action pairs of a trajectory, are also random variables that appear at each timestep. Typically, we are largely concerned with only one of these random variables, attributed to the first timestep of the trajectory $Z(\tau) \triangleq Z_1$, since returns are computed in hindsight. Naturally, of course, there is an entire sequence of these return random variables $Z_1,\ldots,Z_H$ at our disposal. Accordingly, a quantity that may be of great interest when contemplating issues of credit assignment in RL is the following:
\begin{align*}
    \mc{I}(\tau,Z_1,\ldots,Z_H) &= \mc{I}(\tau_1,\ldots,\tau_H;Z_1,\ldots,Z_H)
\end{align*}

which captures all information between a completed trajectory and the sequence of observed returns at each timestep. Recall the chain rule of mutual information:
\begin{align*}
    \mc{I}(X_1,\ldots,X_n;Y) &= \sum\limits_{i=1}^n \mc{I}(X_i;Y|X^{i-1})
\end{align*}
 
 where $X^{i-1} = (X_1,\ldots,X_{i-1})$ and $X^{-1} = \emptyset$. In the previous section, this allowed for a decomposition of the trajectory in temporal order so that we could examine a current state-action pair $\tau_h$ conditioned on the history $\tau^{h-1}$. The analogous step for the sequence of return variables creates a slight oddity where we have the return at a timestep $Z_h$ conditioned on the returns of previous timesteps $Z^{h-1}$. Here, the temporal ordering that was advantageous in breaking apart a trajectory now results in conditioning on returns that are always computed after observing $Z_h$. Fortunately, multivariate mutual information is not sensitive to any particular ordering of the random variables, a fact which can be demonstrated quickly for the two-variable case:
 \begin{fact}
 Let $X,Y,Z$ be three random variables.
\begin{align*}
    \mc{I}(X;Y,Z) &= \mc{I}(X;Z,Y)
\end{align*}
 \label{fact:mi_swap} 
 \end{fact}
 
 Fact \ref{fact:mi_swap} implies that we have a choice between a forward view (for processing variables in temporal order) and a backwards view (for processing in hindsight). This fact by itself is an interesting property of information theory that may deserve more attention in its own right as it blurs the line between the forward-looking perspective of RL and the opposing retrospective view used by credit-assignment techniques for supervised learning~\citep{ke2018sparse}. It is also reminiscent of the forward and backwards views of the widely-studied eligibility trace~\citep{sutton1985temporal,sutton1988learning,sutton1998introduction,singh1996reinforcement}. Since we would like to consider the impact of the entire trajectory on each individual return, we can begin by decomposing the return random variables in hindsight:
\begin{align*}
    \mc{I}(\tau_1,\ldots,\tau_H;Z_1,\ldots,Z_H) &= \sum\limits_{h=1}^H \mc{I}(Z_h; \tau_1,\ldots,\tau_H|Z_{h+1}^H)
\end{align*}
 
 Further expansion of the above multivariate mutual information gives rise to the following proposition that draws a direct connection to causal information theory.
 \begin{proposition}
 Let $\tau = (\tau_1,\ldots,\tau_H)$ be a $H$-step trajectory and let $Z^H = (Z_H,Z_{H-1},\ldots,Z_1)$ be the associated time-sychronized sequence of return random variables. Then, we have that $$ \mc{I}(\tau;Z_1,\ldots,Z_H) = \mc{I}(\tau^H \ra Z^H)$$
 \label{prop:directedinfo}
 The proof is provided in Appendix~\ref{sec:proofs_causal}.
 \end{proposition}
 
 Notice that, in general, Proposition \ref{prop:directedinfo} is not always true for two arbitrary, time-synchronized stochastic processes as the multivariate mutual information and directed information obey a conservation law~\citep{massey2005conservation}. 
 
 By following the first steps from the proof of Proposition \ref{prop:directedinfo}, we can also recover an analog to Proposition \ref{prop:credit} that prescribes a connection between $\mc{I}(\tau;Z_1,\ldots,Z_H)$ and individual rewards $R_h$. 
 
 \begin{proposition}
 \begin{align*}
     \mc{I}(\tau;Z_1,\ldots,Z_H) &= \mc{I}(\tau^H \ra Z^H) = \sum\limits_{h=1}^H \mc{H}(R_h |Z_{h+1}^H)
 \end{align*}
 The proof is provided in Appendix~\ref{sec:proofs_causal}.
 \label{prop:directed_credit}
 \end{proposition}
 
 Taken together, Propositions \ref{prop:directedinfo} and \ref{prop:directed_credit} offer an interesting connection between information theory and causal inference, a link which has appeared before~\citep{amblard2013relation}. We leave the question of how an agent might leverage such quantities to actively reason about the underlying causal structure of the environment to future work.
 
\section{Discussion \& Conclusion}
\label{sec:disc}

In this work, we take an initial step towards a rigorous formulation of the credit-assignment problem in reinforcement learning. At the core of our approach is information theory, which we find naturally suited for obtaining quantitative answers to the core question facing an agent when dealing with credit assignment: \textit{how does choosing an action $a$ in state $s$ affect future return?} While this work offers preliminary ideas for how information theory can then be used to to understand credit assignment, it remains to be seen how these measures can be integrated into existing RL algorithms.

% --- Bibliography ---
\bibliographystyle{plainnat}
\bibliography{references}
\newpage

\appendix

\section{Background}

In this section, we review standard quantities in information theory as well as causal information theory. For more background on information theory, see \citet{cover2012elements}.

\subsection{Information Theory}

\ddef{Entropy \& Conditional Entropy}{
For a discrete random variable $X$ with density function $p(x)$ and support $supp(p(x)) = \mc{X}$, the entropy of $X$ is given by
\begin{align*}
    \mc{H}(X) &= -\bE_{p(x)}[\log(p(x))] \\
    &= - \sum\limits_{x \in \mc{X}} p(x)\log(p(x))
\end{align*}

Similarly, the conditional entropy of a discrete random variable $Y$ given $X$ with density $p(y)$ ($supp(p(y)) = \mc{Y})$ and joint density $p(x,y)$ is given by
\begin{align*}
        \mc{H}(Y|X) &= -\bE_{p(x,y)}[\log(p(y|x))] \\
    &= - \sum\limits_{x \in \mc{X}} \sum\limits_{y \in \mc{Y}} p(x,y)\log(p(y|x))
\end{align*}
}

\ddef{Chain Rule for Entropy}{
For a collection of random variables $X_1,\ldots,X_n$, the joint entropy can be decomposed as a sum of conditional entropies:
\begin{align*}
    \mc{H}(X_1,\ldots,X_n) &= \sum\limits_{i=1}^n \mc{H}(X_i|X_1,\ldots,X_{i-1})
\end{align*}
}

\ddef{Kullback-Leibler Divergence}{
The KL-divergence between two probability distributions $p,q$ with identical support $\mc{X}$ is
\begin{align*}
    \kl{p}{q} &= \sum\limits_{x \in \mc{X}} p(x)\log(\frac{p(x)}{q(x)})
\end{align*}
}

\ddef{Mutual Information}{
The mutual information between two random variables $X$ and $Y$ is given by
\begin{align*}
    \mc{I}(X;Y) &= \kl{p(x,y)}{p(x)p(y)}\\
    &= \mc{H}(X) - \mc{H}(X | Y) \\
    &= \mc{H}(Y) - \mc{H}(Y | X)
\end{align*}

Also, recognizing that $p(x|y) = \frac{p(x,y)}{p(y)}$, we have that
\begin{align*}
    \mc{I}(X;Y) &= \bE_Y[\kl{p(x|y)}{p(x)}]
\end{align*}
}

\ddef{Conditional Mutual Information}{
The conditional mutual information between two variables $X$ and $Y$, given a third random variable $Z$ is
\begin{align*}
    \mc{I}(X;Y|Z) &= \bE_Z[\kl{p(x,y|z)}{p(x|z)p(y|z)}]\\
    &= \mc{H}(X|Z) - \mc{H}(X | Y,Z) \\
    &= \mc{H}(Y|Z) - \mc{H}(Y | X,Z) \\
    &= \mc{I}(X;Y,Z) - I(X;Z)
\end{align*}

Also, recognizing that $p(x|y, z) = \frac{p(x,y|z)}{p(y|z)}$, we have that
\begin{align*}
    \mc{I}(X;Y|Z) &= \bE_Z[\bE_{Y|Z}[\kl{p(x|y,z)}{p(x|z)}]]
\end{align*}
}

\ddef{Multivariate Mutual Information}{
The multivariate mutual information between three random variables $X$, $Y$, and $Z$ is given by:
\begin{align*}
    \mc{I}(X;Y,Z) &= \mc{I}(X; Z) + \mc{I}(X; Y | Z)\\
\end{align*}
}

\ddef{Chain Rule for Mutual Information}{
For a random variables $X$ and $Z_1,\ldots,Z_n$ the multivariate mutual information decomposes into a sum of conditional mutual information terms:
\begin{align*}
    \mc{I}(X;Z_1,\ldots,Z_n) &= \sum\limits_{i=1}^n \mc{I}(X;Z_i|Z_1,\ldots,Z_{i-1})
\end{align*}
}

\subsection{Causal Information Theory}

\ddef{Causal Conditioning \& Entropy ~\citep{kramer1998directed}}{
Consider two time-synchronized stochastic processes $X^T = (X_1,\ldots,X_T)$ and $Y^T = (Y_1, \ldots, Y_T)$. Let $x^T$ and $y^T$ denote two realizations of the respective processes. The causally conditioned probability of the sequence $y^T$ given $x^T$ is
\begin{align*}
    p(y^T || x^T) &= \prod\limits_{t=1}^T p(y_t | y^{t-1}, x^{t-1}) \\
    &= \prod\limits_{t=1}^T p(y_t | y_1, \ldots, y_{t-1}, x_1, \ldots, x_{t-1})
\end{align*}

The causual entropy of $Y^T$ given $X^T$ is
\begin{align*}
    \mc{H}(Y^T || X^T) &= - \bE_{p(x^T,y^T)}[ \log(p(y^T || x^T)] \\
    &= \sum\limits_{t=1}^T \mc{H}(Y_t | Y^{t-1}, X^{t-1})
\end{align*}
}

\ddef{Directed Information Flow~\citep{massey1990causality,permuter2008directed}}{
Let $X^T = (X_1,\ldots,X_T)$ and $Y^T = (Y_1, \ldots, Y_T)$ denote two time-synchronized stochastic processes. The directed information that flows from $X^T$ to $Y^T$ is given by
\begin{align*}
    \mc{I}(X^T \rightarrow Y^T) &= \sum\limits_{t=1}^T \mc{I}(X^{t}; Y_t | Y^{t-1}) \\
    &= \mc{H}(Y^T) - \mc{H}(Y^T || X^T)
\end{align*}
}
\begin{remark}
Notice that, in general, $\mc{I}(X \rightarrow Y) \neq \mc{I}(Y \rightarrow X)$
\end{remark}

\section{Related Work}
\label{sec:related}

% \citet{harutyunyan2019hindsight} offer a refactoring of the Bellman equation via importance sampling 
While there are various methods for circumventing the challenges of sparse rewards, there are relatively fewer works that directly study the credit assignment problem in RL. Perhaps foremost among them are the classic eligibility traces~\citep{sutton1985temporal,sutton1988learning,sutton1998introduction,singh1996reinforcement,precup2000eligibility,van2020expected} which leverage temporal recency as a heuristic for assigning credit to visited states/state-action pairs during temporal-difference learning. \citet{konidaris2011td_gamma,thomas2015policy} offer a formal derivation of standard eligibility traces, underscoring how the underlying theoretical assumptions on the random variables denoting $n$-step returns are typically not realized in practice. In a similar spirit, our work is also concerned with rectifying the shortcomings of eligibility traces, highlighting how temporal recency acts a poor heuristic in hard credit-assignment problems where informative feedback is scarce and credit must be allocated across a number of timesteps that may far exceed the effective horizon of the trace.

Other recent works offer means of modulating or re-using data collected within the environment to help alleviate the burdens of credit assignment. \citet{arjona2019rudder} give a formal characterization of return-equivalent decision-making problems and introduce the idea of reward re-distribution for dispersing sparse, informative feedback backwards in time to the facilitating state-action pairs while still preserving the optimal policy. They realize a practical instantiation of this RUDDER idea through recurrent neural networks and examining differences between return predictions conditioned on partial trajectories. \citet{harutyunyan2019hindsight} offer a refactoring of the Bellman equation via importance sampling to introduce a hindsight distribution, $h(a|s,Z(\tau))$, over actions conditioned on state and the return $Z(\tau)$ of a completed trajectory $\tau$. This distribution is then used solely for the purposes of hindsight credit assignment (HCA) in determining the impact of a particular action $a$ on an observe outcome $Z(\tau)$ through the likelihood ratio $\frac{h(a|s,Z(\tau))}{\pi(a|s)}$. In our work, we introduce two quantities to help measure the credit of a single behavior step towards an observed return. One of these measures, similar to RUDDER, prescribes examining the information contained in partial trajectories over next rewards (rather than returns). The other measure naturally yields the same likelihood ratio of HCA purely from information theory.

For many years, work at the intersection of information theory and reinforcement learning has been a topic of great interest. Perhaps most recently, there has been a resurgence of interest in the control-as-inference framework~\citep{todorov2007linearly,toussaint2009robot,kappen2012optimal,levine2018reinforcement} leading to maximum-entropy (deep) RL approaches~\citep{ziebart2010modeling,fox2016taming,haarnoja2017reinforcement,haarnoja2018soft}. These methods take the uniform distribution as an uninformative prior over actions and can be more broadly categorized as KL-regularized RL~\citep{todorov2007linearly,galashov2019information,tirumala2019exploiting}. Curiously, these information-theoretic RL objectives, derived from the perspective of probabilistic inference, can also be derived from information-theoretic characterizations of bounded rationality~\citep{tishby2011information,ortega2011information,rubin2012trading,ortega2013thermodynamics}. In this setting, a policy is viewed as a channel in the information-theoretic sense and, as with any channel, there is a cost to channel communication (mapping individual states to actions) that goes unaccounted for in the standard RL objective of reward maximization. By modeling this communication cost explicitly, these works also arrive at the KL-regularized RL objective. Examining the communication rate over time, rather than the instantaneous cost, yields a natural analog to these approaches~\citep{tiomkin2017unified} articulated in terms of causal information theory~\citep{kramer1998directed}. Orthogonally, \citet{russo2016information,russo2018learning} study the role of information theory in analyzing efficient exploration; they introduce the information ratio to characterize the balance between taking regret-minimizing actions (exploitation) and actions that have high information gain (exploration), leading to a general regret bound for the multi-armed bandit setting. Empirical work studying effective heuristics for guiding exploration in deep RL have also made great use of information-theoretic quantities~\citep{houthooft2016vime,kim2019emi}. Tackling the issue of generalization in RL, \citep{abel2019state} use rate-distortion theory~\citep{shannon1959coding,berger1971rate} to reformulate the learning of state abstractions~\citep{li2006towards} as as optimization problem in the space of lossy-compression schemes.

A common thread among all the aforementioned work on information theory and RL is that the quantities leveraged are exclusively concerned with only the states and/or actions taken by an agent. In contrast, the information-theoretic quantities explored in this work also incorporate a focus on returns. Given the recent successes of distributional RL~\citep{sobel1982variance,chung1987discounted,bellemare2017distributional,hessel2018rainbow} that explicitly draw attention to the return random variable (and its underlying distribution) as a quantity of interest, it seems natural to re-examine the role that information theory might play in RL with this critical random variable involved. Moreover, a natural place to explore such a connection is the credit assignment problem which directly asks about the dependence (or information) that individual steps of behavior carry about returns (outcomes).

Coincidentally, there are numerous works in the information-theory community which study MDPs in the context of computing and optimizing multivariate mutual information and directed information~\citep{tatikonda2005markov,tatikonda2008capacity,permuter2008capacity,li2018information}. These approaches formulate a specific MDP and applying dynamic programming to recover channel capacity as the corresponding optimal value function. There have also been works examining the success of neural networks for computing and optimizing directed information~\citep{aharoni2020capacity,aharoni2020reinforcement}. While our work moves in the opposite direction to ask how information theory can help address a core challenge of RL, we may turn to these approaches for inspiration on how to employ our information-theoretic quantities for credit assignment in practice.

\section{Proofs: Information-Theoretic Credit Assignment}
\label{sec:proofs}

Here we present the full version of all theoretical results presented in the main paper.

\subsection{Measuring Credit}
\label{sec:proofs}
\begin{fact}
Let $X,Y$ be two discrete random variables and define $S = X + Y$. Then
\begin{align*}
    \mc{H}(S|X) = \mc{H}(Y|X)
\end{align*}
\begin{dproof}
\begin{align*}
    \mc{H}(S | X) &= -\bE_x[\sum\limits_s p(S = s | X = x) \log(p(S = s | X = x))] \\
    &= -\bE_x[\sum\limits_s p(Y = s-x | X = x) \log(p(Y = s-x | X = x))] \\
    &= -\bE_x[\sum\limits_y p(Y = y | X = x) \log(p(Y = y | X = x))] \\ 
    &= \mc{H}(Y|X)
\end{align*}
where we perform a change of variables $y=s-x$ in the third line
\end{dproof}
\label{fact:centropy}
\end{fact}

\begin{proposition}
Let $\pi$ be a fixed behavior policy such that $\tau \sim \rho^\pi$. Let $R_h$ be a random variable denoting the reward observed at timestep $h$ (where the randomness of the deterministic reward follows from the randomness of the state-action pair at $h$, $\tau_h$). It follows that:
\begin{align*}
    \mc{I}\left(Z(\tau);\tau_h|\tau^{-h}\right) &= \mc{H}\left(R_h | \tau^{h-1}\right)
\end{align*}
\begin{dproof}

\begin{align*}
    \mc{I}\left(Z(\tau);\tau_h|\tau^{-h}\right) &\stackrel{(a)}{=} \mc{I}\left(Z(\tau);\tau_h,\tau^{-h}\right) - \mc{I}\left(Z(\tau);\tau^{-h}\right) \\
    &= \mc{I}\left(Z(\tau);\tau\right) - \mc{I}\left(Z(\tau);\tau^{-h}\right) \\
    &\stackrel{(b)}{=} \mc{H}\left(Z(\tau)\right) - \mc{H}\left(Z(\tau)|\tau\right) - \mc{H}\left(Z(\tau)\right) + \mc{H}\left(Z(\tau)|\tau^{-h}\right) \\
    &= \mc{H}\left(Z(\tau)|\tau^{-h}\right) - \mc{H}\left(Z(\tau)|\tau\right) \\
    &\stackrel{(c)}{=} \mc{H}\left(Z(\tau)|\tau^{-h}\right) \\
    &\stackrel{(d)}{=} \mc{H}\left(R_h | \tau^{-h}\right) \\
    &= \mc{H}\left(R_h | \tau^{h-1}, \tau_{h+1}^H\right) \\
    &\stackrel{(e)}{=} \mc{H}\left(R_h | \tau^{h-1}\right)
\end{align*}
where the steps follow from: $(a)$ the chain rule of mutual information, $(b)$ the definition of mutual information, $(c)$ $Z(\tau)$ is a deterministic function of $\tau$ under a deterministic reward function, $(d)$ applies Fact \ref{fact:centropy} on $Z(\tau) = \sum\limits_{h=1}^H \gamma^{h-1} R_h$, and $(e)$ $R_h$ is independent of the future trajectory $\tau_{h+1}^H$.
\end{dproof}
% \label{prop:credit}
\end{proposition}

\subsection{Hindsight}
\label{sec:hingsight_proof}

\begin{proposition}
Let $\pi$ be a fixed behavior policy such that $\tau \sim \rho^\pi$ and let $h(a|s,Z(\tau))$ be the hindsight distribution as defined above. We have that
\begin{align*}
    \mc{I}(Z(\tau);\tau_h|\tau^{h-1}) &= \bE_{\tau^h}\left[\bE_{Z(\tau) | \tau^h}\left[ \log\left(\frac{h(a_h|s_h, Z(\tau))}{\pi(a_h | s_h)}\right)\right] \right]
\end{align*}

Moreover,
\begin{align*}
    \mc{I}(Z(\tau);\tau) &= \sum\limits_{h=1}^H \bE_{\tau^h}\left[\bE_{Z(\tau) | \tau^h}\left[ \log\left(\frac{h(a_h|s_h, Z(\tau))}{\pi(a_h | s_h)}\right)\right] \right]
\end{align*}
% \label{prop:hca}
\begin{dproof}
Notice that by the definition of conditional mutual information:
\begin{align*}
    \mc{I}(Z(\tau);\tau_h|\tau^{h-1}) &= \bE_{\tau^{h-1}}\left[\bE_{\tau_h | \tau^{h-1}}\left[\kl{p(Z(\tau) | \tau_h, \tau^{h-1})}{p(Z(\tau) | \tau^{h-1})}\right]\right] \\
    &= \bE_{\tau^h}\left[\kl{p(Z(\tau) | \tau^h)}{p(Z(\tau) | \tau^{h-1})} \right] \\
    &= \bE_{\tau^h}\left[\bE_{Z(\tau) | \tau^h}\left[ \log\left(\frac{p(Z(\tau) | \tau^h)}{p(Z(\tau) | \tau^{h-1})}\right)\right] \right]
\end{align*}
By applying Bayes' rule twice, we have:
\begin{align*}
    \mc{I}(Z(\tau);\tau_h|\tau^{h-1}) &= \bE_{\tau^h}\left[\bE_{Z(\tau) | \tau^h}\left[ \log\left(\frac{p(Z(\tau) | \tau^h)}{p(Z(\tau) | \tau^{h-1})}\right)\right] \right] \\
    &= \bE_{\tau^h}\left[\bE_{Z(\tau) | \tau^h}\left[ \log\left(\frac{p(\tau^h|Z(\tau))p(\tau^{h-1})}{p(\tau^{h-1} | Z(\tau))p(\tau^h))}\right)\right] \right] \\
    &= \bE_{\tau^h}\left[\bE_{Z(\tau) | \tau^h}\left[ \log\left(\frac{h(a_h|s_h, Z(\tau))}{\pi(a_h | s_h)}\right)\right] \right]
\end{align*}

where the final equation follows from the fact that
\begin{align*}
    p(\tau^h) &= \beta(s_1) \pi(a_1|s_1) \prod\limits_{h'=2}^h \pi(a_{h'}|s_{h'}) \mc{T}(s_{h'} | s_{h'-1}, a_{h'-1}) \\
    p(\tau^h | Z(\tau)) &= \beta(s_1) h(a_1|s_1, Z(\tau)) \prod\limits_{h'=2}^h h(a_{h'}|s_{h'},Z(\tau)) \mc{T}(s_{h'} | s_{h'-1}, a_{h'-1}) \\
\end{align*}
with analogous distributions for $p(\tau^{h-1})$ and $p(\tau^{h-1}|Z(\tau))$.

To show the second claim, we simply apply the chain rule of mutual information:
\begin{align*}
    \mc{I}(Z(\tau);\tau) &= \mc{I}(Z(\tau); \tau_1,\tau_2,\ldots,\tau_H) \\
    &= \sum\limits_{h=1}^H \mc{I}(Z(\tau); \tau_h | \tau^{h-1}) \\
    &= \sum\limits_{h=1}^H \bE_{\tau^h}\left[\bE_{Z(\tau) | \tau^h}\left[ \log\left(\frac{h(a_h|s_h, Z(\tau))}{\pi(a_h | s_h)}\right)\right] \right]
\end{align*}
\end{dproof}
\end{proposition}

\subsection{Causal Information Theory \& Hindsight}
\label{sec:proofs_causal}
 \begin{fact}
 Let $X,Y,Z$ be three random variables.
 \begin{align*}
     \mc{I}(X;Y,Z) &= \mc{I}(X;Z,Y)
 \end{align*}
 \begin{dproof}
  \begin{align*}
     \mc{I}(X;Y,Z) &= \mc{I}(X;Z) + \mc{I}(X;Y|Z) \\
     &= \mc{H}(X) - \mc{H}(X|Z) + \mc{H}(X|Z) - \mc{H}(X|Y,Z) \\
     &= \mc{H}(X) - \mc{H}(X|Y,Z) \\
     &= \mc{H}(X) - \mc{H}(X|Y) + \mc{H}(X|Y) - \mc{H}(X|Y,Z) \\
     &= \mc{I}(X;Y) + \mc{I}(X;Z|Y) \\
     &= \mc{I}(X;Z,Y)
 \end{align*}
 \end{dproof}
%  \label{fact:mi_swap}
 \end{fact}
 
 \begin{proposition}
 Let $\tau = (\tau_1,\ldots,\tau_H)$ be a $H$-step trajectory and let $Z^H = (Z_H,Z_{H-1},\ldots,Z_1)$ be the associated time-sychronized sequence of return random variables. Then, we have that
 \begin{align*}
     \mc{I}(\tau;Z_1,\ldots,Z_H) &= \mc{I}(\tau^H \ra Z^H)
 \end{align*}
 \begin{dproof}
 Notice that, for any timestep $h$, we can group the state-action pairs contained in a trajectory by $h$ into a past $\tau^{h-1} = (\tau_1,\ldots,\tau_{h-1})$, present $\tau_h$, and future $\tau_{h+1}^H = (\tau_{h+1},\ldots,\tau_H)$.
  \begin{align*}
     \mc{I}(\tau;Z_1,\ldots,Z_H) &= \sum\limits_{h=1}^H \mc{I}(Z_h; \tau_1,\ldots,\tau_H|Z_{h+1}^H) \\
     &= \sum\limits_{h=1}^H \mc{I}(Z_h; \tau^{h-1}, \tau_h, \tau_{h+1}^H |Z_{h+1}^H) \\
     &= \sum\limits_{h=1}^H \mc{I}(Z_h; \tau^{h-1}, |Z_{h+1}^H) + \mc{I}(Z_h; \tau_h, |\tau^{h-1}, Z_{h+1}^H) + \mc{I}(Z_h; \tau_{h+1}^H, |\tau^{h-1}, \tau_h, Z_{h+1}^H) 
 \end{align*}
 Recall that we take the reward function of our MDP to be deterministic. Consequently, we have that $Z_h$ is a deterministic function of $\tau_h$ and $Z_{h+1}$. Thus, $\mc{I}(Z_h;\tau_{h+1}^H|\tau^{h-1},\tau_h,Z_{h+1}^H) = 0$ and we're left with
 \begin{align*}
     \mc{I}(\tau;Z_1,\ldots,Z_H) &= \sum\limits_{h=1}^H \mc{I}(Z_h; \tau^{h-1}, |Z_{h+1}^H) + \mc{I}(Z_h; \tau_h, |\tau^{h-1}, Z_{h+1}^H) \\
     &= \sum\limits_{h=1}^H \mc{I}(Z_h; \tau^{h-1}, \tau_h |Z_{h+1}^H) \\
     &= \sum\limits_{h=1}^H \mc{I}(Z_h; \tau^h |Z_{h+1}^H) \\
     &= \mc{I}(\tau^H \ra Z^H)
 \end{align*}
 \end{dproof}
%  \label{prop:directedinfo}
 \end{proposition}
 
 \begin{fact}
 Let $X,Y,Z$ be three discrete random variables and let $S = X + Y$. Then,
 \begin{align*}
     \mc{I}(S;Y|Z) &= \mc{I}(X;Y|Z)
 \end{align*}
 \begin{dproof}
  \begin{align*}
     \mc{I}(S;Y|Z) &= \mc{I}(X+Z;Y|Z) \\
     &= \mc{H}(X+Z|Z) - \mc{H}(X+Z|Y,Z) \\
     &= \mc{H}(X|Z) - \mc{H}(X|Y,Z)\\
     &= \mc{I}(X;Y|Z)
 \end{align*}
 \end{dproof}
 where the third line applies Fact \ref{fact:centropy}.
 \label{fact:summi}
 \end{fact}
 
 \begin{proposition}
 \begin{align*}
     \mc{I}(\tau;Z_1,\ldots,Z_H) &= \mc{I}(\tau^H \ra Z^H) = \sum\limits_{h=1}^H \mc{H}(R_h |Z_{h+1}^H)
 \end{align*}
 \begin{dproof}
 Using $\tau^{h-1}$ to denote the past trajectory from a given timestep as above
 \begin{align*}
     \mc{I}(\tau;Z^H) &= \sum\limits_{h=1}^H \mc{I}(Z_h; \tau^{h-1}, |Z_{h+1}^H) + \mc{I}(Z_h; \tau_h, |\tau^{h-1}, Z_{h+1}^H) \\ 
     &= \sum\limits_{h=1}^H \mc{I}(R_h; \tau^{h-1}|Z_{h+1}^H) + \mc{I}(R_h; \tau_h |\tau^{h-1}, Z_{h+1}^H) \\
     &= \sum\limits_{h=1}^H \mc{I}(R_h; \tau^{h-1}, \tau_h |Z_{h+1}^H) \\
     &= \sum\limits_{h=1}^H \mc{I}(R_h; \tau^h |Z_{h+1}^H) \\
     &= \sum\limits_{h=1}^H \mc{H}(R_h |Z_{h+1}^H) - \mc{H}(R_h |\tau^h, Z_{h+1}^H) \\
     &= \sum\limits_{h=1}^H \mc{H}(R_h |Z_{h+1}^H)
 \end{align*}
 where the second line recognizes that $Z_h = R_h + \gamma Z_{h+1}$ and employs Fact \ref{fact:summi}. The last line follows from the fact that $R_h$ is a deterministic function of $\tau_h$.
 \end{dproof}
 \end{proposition}

\end{document}

%% file: commands.tex
\definecolor{dblue}{RGB}{98, 140, 190}
\definecolor{dlblue}{RGB}{216, 235, 255}
\definecolor{dgreen}{RGB}{124, 155, 127}
\definecolor{dpink}{RGB}{207, 166, 208}
\definecolor{dyellow}{RGB}{255, 248, 199}
\definecolor{dgray}{RGB}{46, 49, 49}

\newcommand{\durl}[1]{\textcolor{dblue}{\underline{\url{#1}}}}

\newcommand{\mc}[1]{\mathcal{#1}}

\newcommand{\bE}{\mathbb{E}}

\newcommand{\bR}{\mathbb{R}}

\newcommand{\kl}[2]{D_{KL}(#1 || #2)}

\newcommand{\ra}{\rightarrow}

\newmdenv[
  topline=false,
  bottomline=false,
  rightline = false,
  leftmargin=10pt,
  rightmargin=0pt,
  innertopmargin=0pt,
  innerbottommargin=0pt
]{innerproof}

\newenvironment{dproof}[1][Proof]{\begin{proof}[#1] \text{\vspace{2mm}} \begin{innerproof}}{\end{innerproof}\end{proof}\vspace{4mm}}

\newcounter{DaveDefCounter}
\newcommand{\ddef}[2]
{
\begin{mdframed}[roundcorner=1pt, backgroundcolor=white]
\vspace{1mm}
{\bf Definition \theDaveDefCounter} (#1): {\it #2}
\stepcounter{DaveDefCounter}
\end{mdframed}
}

\newtheorem{proposition}{Proposition}
\newtheorem{remark}{Remark}

\newtheorem{fact}{Fact}